\documentclass[runningheads]{llncs}
\usepackage{graphicx}
\usepackage{comment}
\usepackage{amsmath,amssymb} 
\usepackage{color}
\usepackage{epsfig}
\usepackage{graphicx}
\usepackage{amsmath}
\usepackage{amssymb}
\usepackage{array}
\usepackage{multirow}

\usepackage[]{algorithm}
\usepackage{algorithmic}

\def\Fig#1{{Fig.~\ref{fig:#1}}}

\def\Table#1{{Table~\ref{tbl:#1}}}
\def\eg{{e.g.}}
\def\etal{{et al.}}
\def\ie{{i.e.}}

\begin{document}
\pagestyle{headings}
\mainmatter
\def\ECCVSubNumber{4369}  

\title{Universal Lesion Detection by Learning from Multiple Heterogeneously Labeled Datasets} 

\titlerunning{Universal Lesion Detection on Multiple Heterogeneous Datasets}
%
\author{Ke Yan \and
Jinzheng Cai \and
Adam P.~Harrison \and
Dakai Jin \and
Jing Xiao \and
Le Lu}
\authorrunning{K.~Yan et al.}
%
\institute{PAII Inc., Bethesda MD 20817, USA }
\maketitle

\begin{abstract}
Lesion detection is an important problem within medical imaging analysis. Most previous work focuses on detecting and segmenting a specialized category of lesions (\eg, lung nodules). However, in clinical practice, radiologists are responsible for finding all possible types of anomalies. The task of universal lesion detection (ULD) was proposed to address this challenge by detecting a large variety of lesions from the whole body. There are multiple heterogeneously labeled datasets with varying label completeness: DeepLesion, the largest dataset of 32,735 annotated lesions of various types, but with even more missing annotation instances; and several fully-labeled single-type lesion datasets, such as LUNA for lung nodules and LiTS for liver tumors. In this work, we propose a novel framework to leverage all these datasets together to improve the performance of ULD. First, we learn a multi-head multi-task lesion detector using all datasets and generate lesion proposals on DeepLesion. Second, missing annotations in DeepLesion are retrieved by a new method of embedding matching that exploits clinical prior knowledge. Last, we discover suspicious but unannotated lesions using knowledge transfer from single-type lesion detectors. 
In this way, reliable positive and negative regions are obtained from partially-labeled and unlabeled images, which are effectively utilized to train ULD. To assess the clinically realistic protocol of 3D volumetric ULD, we fully annotated 1071 CT sub-volumes in DeepLesion. Our method outperforms the current state-of-the-art approach by 29\% in the metric of average sensitivity. 

\end{abstract}

\section{Introduction}
\label{sec:intro}

At the core of oncology imaging for diagnosis of potential cancers, radiologists are responsible to find and report all possible abnormal findings (e.g., tumors, lymph nodes, and other lesions). It is not only time-consuming to scan through a 3D medical image, human readers may also miss some abnormal findings. This spurs research on automated lesion detection to decrease reading time and improve accuracy~\cite{Litjens2017survey,Sahiner2018survey}. Existing work commonly focus on lesions of specific types and organs. For example, lung nodules~\cite{Liao2017noisyor,Zhu2018DeepEM,dou2017multilevel,Ding2017nodule}, liver tumors~\cite{Wang2019LiTS,Li2018HDenseUNet,Cohen2016liver}, and lymph nodes~\cite{Bouget2019lymph,Roth2016randView,Liu2016lymph,Shin2016TMICNN} have been extensively studied. However, in clinical scenarios, a CT scan may contain multiple types of lesions in different organs. For instance, metastasis (\eg, lung cancer) can spread to regional lymph nodes and other body parts (\eg, liver, bone, adrenal, etc.). To help radiologists find all of them, a universal lesion detection (ULD) algorithm, which can identify a variety of lesions in the whole body, is ideal. Designing a model for each organ / lesion type is inefficient and less scalable, significantly increasing inference time and model size. For rare lesion types with fewer training data, single-type models have higher risks of overfitting. More importantly, given the wide range of lesion types, a group of single-type models will still miss some infrequent types. Hence, a ULD system that covers all kinds of lesions is of great clinical value, approaching to address radiologists' daily workflows and real needs. 

To learn an effective ULD system, a comprehensive and diverse dataset of lesion images is required. The conventional data curation paradigm demands experienced radiologists to relabel all lesions, thus is difficult to acquire. Most manually-labeled lesion datasets~\cite{Setio2017LUNA,Bilic2019LiTS,NIH_LN_dataset} are relatively small ($ \sim $1K lesions) and contain specific single lesion types. To tackle this problem, the DeepLesion dataset~\cite{Yan2018graph,Yan2018DeepLesion} was collected by mining lesions directly from the picture archiving and communication system (PACS), which stores the RECIST~\cite{Eisenhauer2009RECIST} markers already annotated by radiologists during their daily work. DeepLesion includes over 32K lesions on various body parts in computed tomography (CT) scans. The ULD task accuracy has been constantly improving~\cite{Wang2019universal,Yan20183DCE,Zlocha2019retina,Li2019MVP,Wang2019LiTS,Yan2019MULAN} upon the release of dataset~\cite{Yan2018DeepLesion}. Along with its large scale and ease of collection, DeepLesion also has a limitation: not all lesions in every image were annotated. This is because radiologists generally mark only representative lesions in each scan \cite{Eisenhauer2009RECIST} in their routine work. This missing annotation or incomplete label problem can also be found in other object detection datasets \cite{Niitani2019sample,Wu2019soft}, which will cause incorrect training signals (some negative proposals are actually positive), 
resulting in lower detection accuracy. In medical images, the appearance of lesions and non-lesions can be quite similar, making it difficult to mine missing annotations and ensure that they are true lesions.

Several public, fully-labeled, and single-type lesion datasets~\cite{Setio2017LUNA,Bilic2019LiTS,NIH_LN_dataset} exist and provide annotations of specific lesion types. For example, the LiTS dataset~\cite{Bilic2019LiTS} contains primary and secondary liver tumors in 201 CT scans. 
While DeepLesion~\cite{Yan2018DeepLesion} is large, universal, but partially-labeled, these datasets~\cite{Setio2017LUNA,Bilic2019LiTS,NIH_LN_dataset} are small, specialized, but fully-labeled. This heterogeneity of dataset labels poses challenges in leveraging multi-source datasets 
in ULD.

\begin{figure}[]
	\centering
	\includegraphics[width=\linewidth,trim=0 110 0 0, clip]{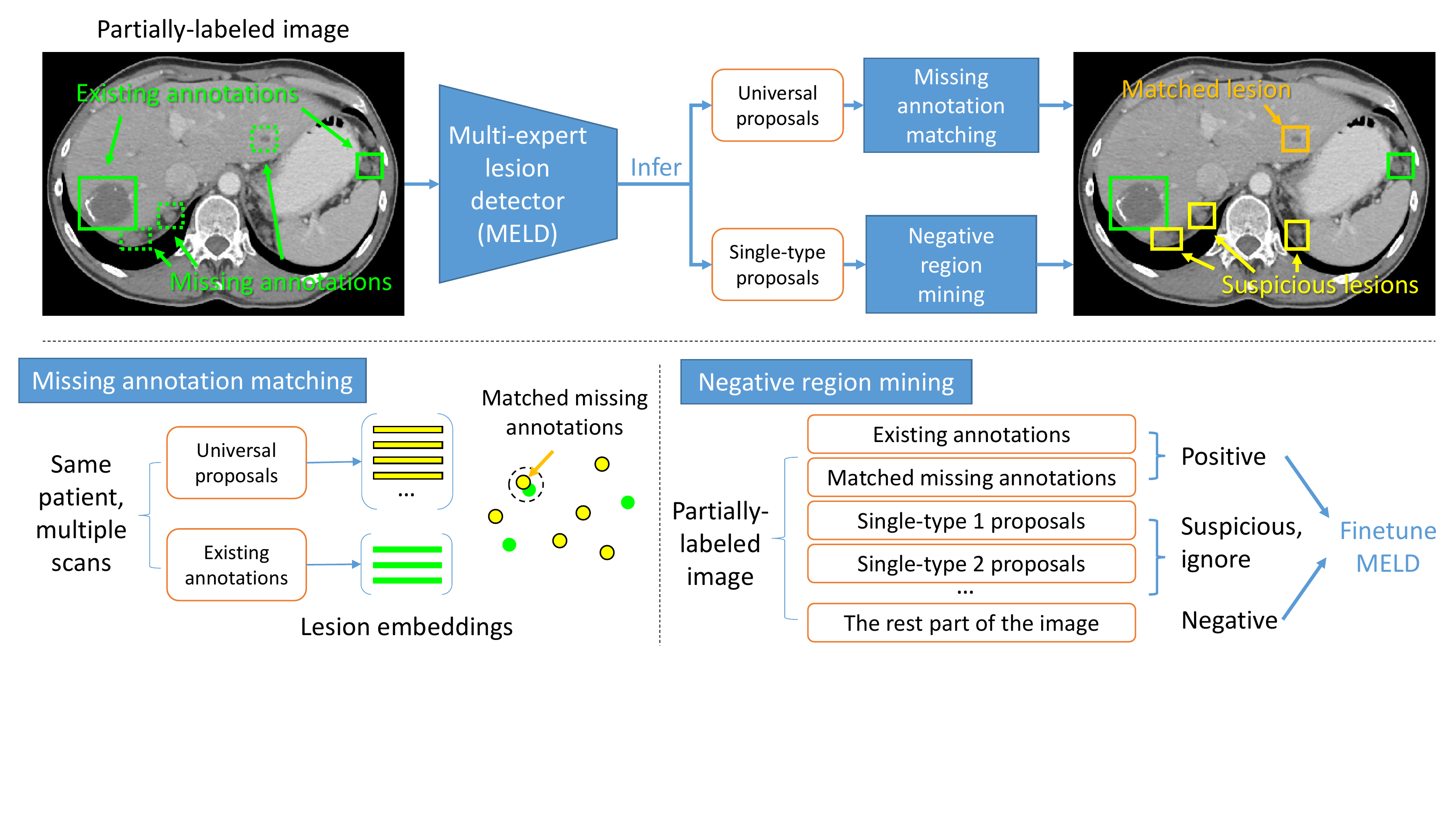} 
	\caption{Overview of the proposed universal lesion detection (ULD) framework. First, we train a multi-expert lesion detector (MELD) using all universal and single-type datasets. Then, we apply MELD on all partially-labeled and unlabeled training images to generate universal and single-type proposals. Next, missing annotation matching (MAM) and negative region mining (NRM) are performed to find positive and negative regions from training images, which are then used to finetune MELD.}
	\label{fig:framework} \vspace{-3mm}
\end{figure}

In this paper, we propose to alleviate the missing annotation problem and leverage multiple datasets to improve ULD accuracy. Our framework is shown in \Fig{framework}. First, we design a lesion detector with several head branches to focus on lesions of different organs. It is trained on multiple lesion datasets in a multi-task fashion, which can handle the heterogeneous label problem. Given a test image, it can predict several groups of lesion proposals matching the semantics of each dataset in \cite{Yan2018DeepLesion,Setio2017LUNA,Bilic2019LiTS,NIH_LN_dataset}. It is named ``multi-expert lesion detector'' (MELD) since each group of proposal is predicted by an ``expert'' learned from one dataset. Then, we employ MELD on the partially-labeled and unlabeled slices in DeepLesion to generate lesion proposals.

\begin{figure}[t]
	\centering
	\includegraphics[width=\linewidth,trim=0 320 0 0, clip]{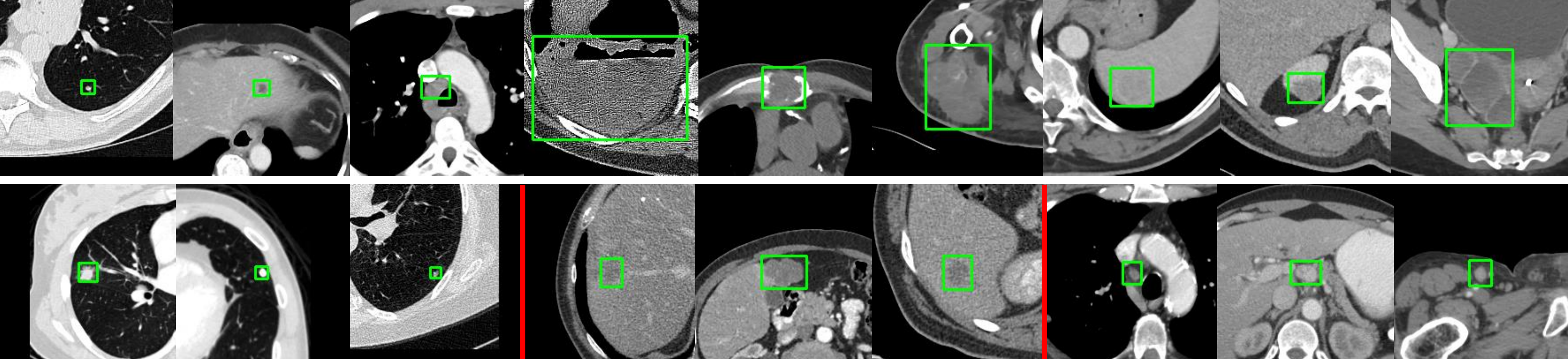} 
	\caption{Exemplar lesions from DeepLesion (first row), LUNA (second row 1--3), LiTS (4--6), and NIH-LN (7--9). DeepLesion has an overlap with the single-type datasets (first row 1--3), but it also includes many clinically significant lesion types that are not covered by other datasets (first row 4--9), demonstrating the value a ULD system.}
	\label{fig:dataset_examples} \vspace{-3mm}
\end{figure}

The key components of our framework are two novel algorithms to mine missing positive and reliable negative regions. In the missing annotation matching (MAM) algorithm, a lesion embedding~\cite{Yan2019Lesa} is extracted for each proposal which describes its body part, type, and attributes. By comparing the embedding of each proposal and each annotated lesion within each patient's multiple CT scans, we can find unannotated lesions that are similar to existing annotations. In the negative region mining (NRM) algorithm, we obtain suspicious lesions from the proposals of the single-type experts in MELD. Because these proposals can be noisy, we do not treat them as lesions, but consider the rest part of the image as reliable negative region without lesions. These positive and negative regions are then used to finetune MELD. We employ three single-type datasets in our framework, namely LUNA (LUng Nodule Analysis)~\cite{Setio2017LUNA}, LiTS (Liver Tumor Segmentation Benchmark)~\cite{Bilic2019LiTS}, and NIH-LN (NIH Lymph Node)~\cite{NIH_LN_dataset}. Notice that it is not our goal to achieve new state-of-the-art results on these specialized datasets. \Fig{dataset_examples} exhibits exemplar lesions from the four datasets. In our experiments, 27K missing annotations and 150K suspicious lesions were found in 233K partially-labeled and unlabeled slices. For evaluation, we manually annotated all lesions in 1K sub-volumes in DeepLesion as the test set\footnote{\scriptsize These annotations will be made publicly available. We were unable to annotate full volumes as images in DeepLesion were released in sub-volumes containing 7$ \sim $220 consecutive slices.}. The original test set in DeepLesion was annotated on selected key slices. It is different from clinical practice where 3D volumetric data are used. Besides, the key-slice test set was not fully annotated, leading to inaccurate performance evaluation. In the fully-labeled volumetric test set, our method outperforms the current state-of-the-art method on DeepLesion by 29\% (average sensitivity from 32.4\% to 41.8\%). 

The main contributions of this paper are fourfold. \textbf{1)} The heterogeneous dataset fusion problem in lesion detection are tackled for the first time via our simple yet effective MELD network; \textbf{2)} We propose two novel methods, \ie~missing annotation matching (MAM) and negative region mining (NRM), to alleviate the missing annotation problem, enabling us to leverage  partially-labeled and unlabeled images in training successfully; \textbf{3)} MELD and NRM can transfer meaningful knowledge from single-type datasets to universal lesion detection models; and \textbf{4)} ULD accuracy on DeepLesion \cite{Yan2018DeepLesion} is significantly improved upon previous state-of-the-art work \cite{Yan2019MULAN}.

\section{Related Work}
\label{sec:rel_work}

{\bf Universal lesion detection:} ULD has been improved using 3D context~\cite{Yan20183DCE,Yan2019MULAN,Wang2019LiTS}, attention mechanism~\cite{Wang2019universal,Zlocha2019retina,Li2019MVP,Wang2019LiTS}, multi-task learning~\cite{Yan2019MULAN,Wang2019universal}, and hard negative mining~\cite{Tang2019Uldor}. 3D context information in neighboring slices is important for detection, as lesions may be less distinguishable in just one 2D axial slice. Volumetric attention~\cite{Wang2019LiTS} exploited 3D information with multi-slice image inputs and a 2.5D network and obtained top results on the LiTS dataset. In~\cite{Zlocha2019retina,Li2019MVP,Wang2019LiTS}, attention mechanisms were applied to emphasize important regions and channels in feature maps. MVP-Net~\cite{Li2019MVP} learned to encode position (body part) information in an attention module. The multi-task universal lesion analysis network (MULAN)~\cite{Yan2019MULAN} achieved the state-of-the-art accuracy on DeepLesion with a 3D feature fusion strategy and the Mask R-CNN~\cite{He2017MaskRCNN} architecture. It jointly learned lesion detection, segmentation, and tagging with a proposed score refinement layer to improve detection with 171 lesion tags. However, it did not handle the missing annotations and the variation of lesions in different organs, neither did it leverage multiple datasets. ULDor~\cite{Tang2019Uldor} mined hard negative proposals with a trained detector to retrain the model, but the mined negatives may actually contain positives because of missing annotations. Inspired by~\cite{Yan2019MULAN,Wang2019LiTS}, we build MELD based on 2.5D Mask R-CNN.

{\bf Learning with incomplete labels:} In detection, knowledge distillation~\cite{Hinton2014distill} can help to find missing annotations. The basic idea is to use the prediction of one model to train another. Predictions from multiple transformations of unlabeled data were merged to generate new training annotations in~\cite{Radosavovic2017data}. Dong \etal~\cite{Dong2019few} progressively generated pseudo-boxes from old models to train new ones. Prior knowledge can also help to infer missing annotations. Wu \etal~\cite{Wu2019soft} argued that a proposal with a small overlap with an existing box is less likely to be a missing annotation. Niitani \etal~\cite{Niitani2019sample} introduced part-aware sampling that assumes an object (car) must contain its parts (tire). Jin \etal~\cite{Jin2018mining} mined hard negative and positive proposals from unlabeled videos based on the prior that object proposals should be continuous across frames. In our framework, we leverage embedding matching and knowledge from multiple specialized datasets to find missing annotations and reliable negative regions.

{\bf Multi-task and multi-dataset learning:} Our problem is related to multi-task learning~\cite{Pan2010transfer} where different tasks are learned jointly, which has been proved beneficial in medical imaging~\cite{Xu2018less,Harouni2018universal,Yan2019MULAN}. Because of the difficulty in data annotation of medical images, it is sometimes required to learn from multiple datasets labeled by different institutes using varying criterion \cite{Tajbakhsh2020imperfect}. Zhou \etal~\cite{Zhou2019organ} and Dmitriev \etal~\cite{Dmitriev2019seg} studied how to learn multi-organ segmentation from single-organ datasets, incorporating priors on organ sizes and dataset-conditioned features, respectively. Cohen \etal~\cite{Cohen2020domain} observed that the same class label had different distribution (concept shift) between multiple chest X-ray datasets and simply training with all datasets is not optimal. The relation and corporation of multiple datasets for lesion detection has not been inspected. The domain-attentive universal detector~\cite{Wang2019universal} used a domain attention module to learn DeepLesion as well as 10 other object detection datasets. Yet, it did not exploit the semantic relation between datasets. Our framework leverages the synergy of lesion datasets both to learn better features and to use their semantic overlaps.

\section{Method}
\label{sec:method}
As shown in \Fig{framework}, we first train a multi-expert lesion detector (MELD) to generate proposals, and then perform missing annotation matching (MAM) and negative region mining (NRM) to find missing positive and reliable negative regions to finetune MELD. Knowledge from single-type datasets is transferred to the universal detector in two ways. First, single-type datasets are jointly trained in MELD to help it learn better feature representation for ULD; Second, in NRM, the single-type experts of MELD help to mine suspicious lesions.

\subsection{Multi-Expert Lesion Detector (MELD) for Heterogeneously Labeled Datasets}
\label{subsec:MELD}

\begin{figure}[t]
	\centering
	\includegraphics[width=.85\linewidth,trim=0 150 0 0, clip]{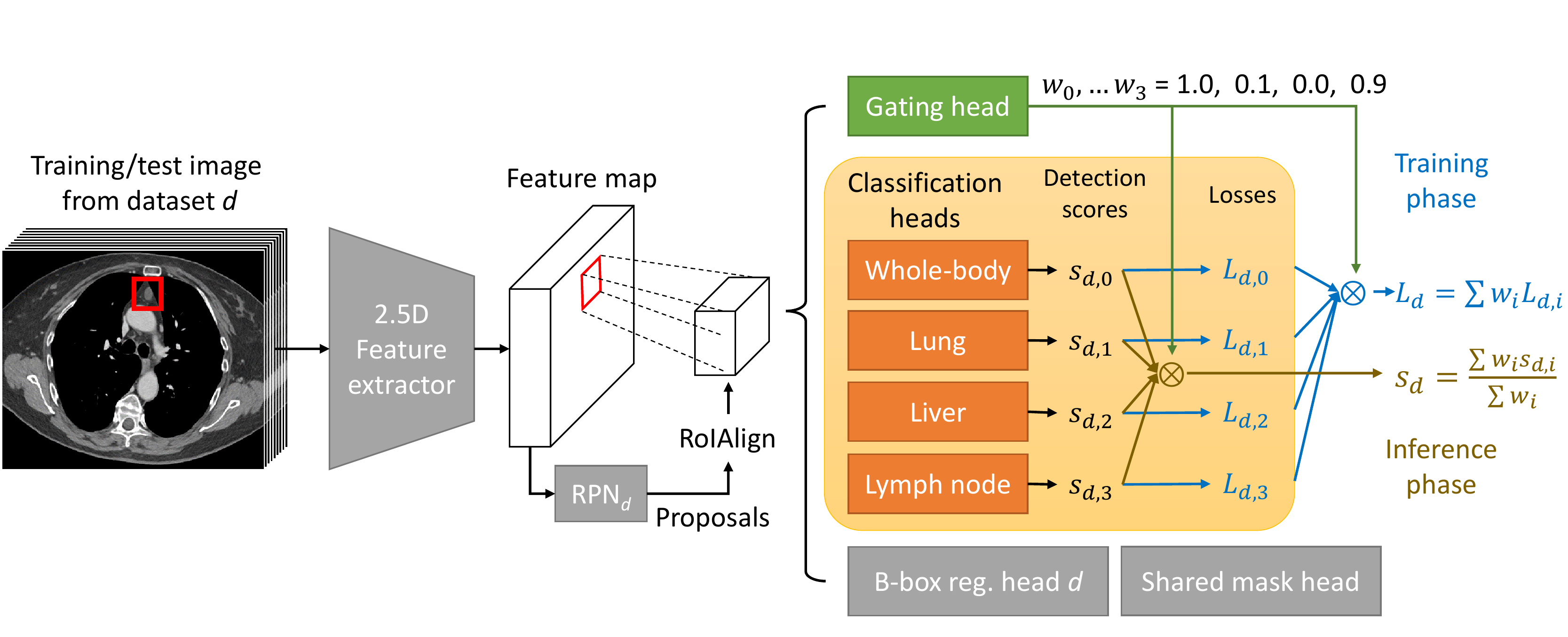} 
	\caption{Framework of the multi-expert lesion detector (MELD). MELD jointly learns multiple lesion datasets in a multi-task fashion. Each dataset $ d $ has its own RPN, detection scores, and bounding-box regression head. MELD also has several classification heads, each focusing on lesions in different organs.}
	\label{fig:MELD_framework} \vspace{-3mm}
\end{figure}

We propose MELD based on an improved Mask R-CNN~\cite{He2017MaskRCNN} architecture. Its overall framework is displayed in \Fig{MELD_framework}. The input of the network is 9 consecutive axial CT slices and the outputs are the detected 2D lesion proposals in the central slice and their segmentation masks. In lesion detection, 3D context in neighboring slices is important, so we use a 2.5D DenseNet backbone plus the feature pyramid network (FPN)~\cite{Lin2016Pyramid} similar to~\cite{Yan2019MULAN}. From the backbone feature, a region proposal network (RPN)~\cite{Ren2015faster} predicts initial lesion proposals and forwards them to the classification, bounding-box regression, and mask heads~\cite{He2017MaskRCNN}. 

In ULD, one challenge is that lesions in different organs have very distinct appearances, while lesions and non-lesions in the same organ can look similar. We propose a divide-and-conquer strategy to alleviate this problem. Existing ULD algorithms treat all kinds of lesions as one class and use a binary classifier to distinguish them from non-lesions. In contrast, MELD has multiple classification heads focusing on lesions of different organs. Lymph node, lung, and liver are the most common organs in DeepLesion~\cite{Yan2019Lesa}, so we build three heads for these three organs and another whole-body head which learns  all lesions. When training, every proposal goes through all heads to obtain detection scores $s_0, \ldots, s_3$ and cross-entropy losses $L_0, \ldots, L_3$. At the same time, we learn a gating head to predict organ weights $w_i \in [0,1], i=0,\ldots,3$, representing how much the proposal belongs to organ $ i $. $ w_0 $ should be always 1 as it corresponds to the whole-body head. The overall loss is $ L=\Sigma_{i=0}^3 w_i L_i$.
When testing, the detection scores are fused by the organ weights and then normalized, i.e., \begin{equation}\label{eq:scorefuse}
s=\left({\Sigma_{i=0}^3 w_i s_i}\right) / \left({\Sigma_{i=0}^3 w_i}\right).
\end{equation}
To train the gating head, we take advantage of the lesion annotation network (LesaNet)~\cite{Yan2019Lesa}. LesaNet was trained on labels mined from radiological reports of DeepLesion. We use it to predict the organ of lesions in DeepLesion, then adopt the predicted scores ($0 \sim 1$) as soft targets to supervise the gating head. This organ stratification strategy allows each classification head to learn organ-specific parameters to model the subtle difference between lesions and non-lesions of the organ. We find it improves the ULD accuracy.

Next, we further extend the above network to jointly learn multiple datasets. In our problem, the datasets are heterogeneously labeled. The definition of lesion in different datasets is \textit{overlapping but not identical}. Single-type datasets lack annotations of other types. For instance, enlarged lymph nodes often exist but were not annotated in LUNA and LiTS. In addition, since the datasets' patient population and collection criteria are all different, there exists a concept shift~\cite{Cohen2020domain,Pan2010transfer}. For example, the distribution of liver tumors in DeepLesion and LiTS may slightly vary. This issue was also found in multiple Chest X-ray datasets~\cite{Cohen2020domain}. Therefore, combining them is not straightforward and it is better to treat different datasets as different learning tasks.

In MELD, we make the datasets share the same network backbone and fully connected layers in the classification heads. 
Each head splits in the last layer and outputs 4 detection scores to match each dataset's semantics. When a training sample comes from dataset $ d $, only the $ d $'th detection score in each classification head will be learned. Additionally, each dataset has its own RPN and bounding-box regression layer~\cite{Wang2019universal}, see \Fig{MELD_framework}. Wang \etal~\cite{Wang2019universal} introduced a domain attention module to learn features from different image domains. Our datasets are from the same image domain (CT scans), so it is feasible to simply make them share features. 
Experiments show that this multi-task strategy improves the accuracy on all datasets. Given a test image, MELD can efficiently predict several groups of lesion proposals matching the semantics of each dataset.

\subsection{Missing Annotation Matching (MAM)}
\label{subsec:MAM}

\begin{figure}[t]
	\centering
	\includegraphics[width=\linewidth,trim=0 300 50 0, clip]{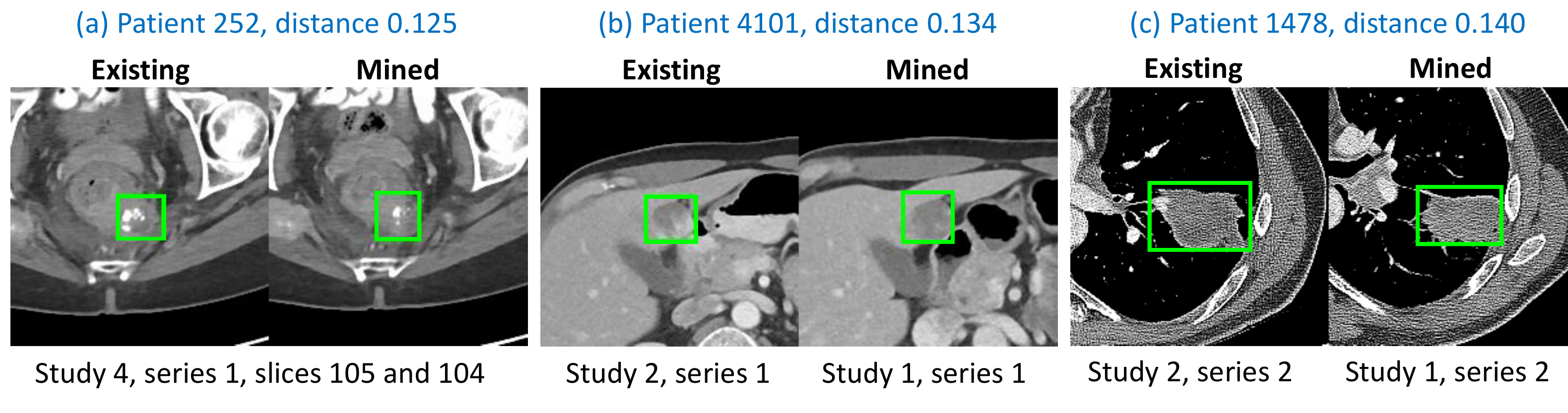} 
	\caption{Examples of matched lesions in DeepLesion. In each sub-plot, the lesion on the left is an existing annotation; the right one is a matched missing annotation in another study/series/slice of the same patient. Their embedding distance is also shown.}
	\label{fig:MAM_example} \vspace{-3mm}
\end{figure}

In clinical practice, each patient generally undergo multiple CT scans (studies) at different time points to monitor their disease progress~\cite{Eisenhauer2009RECIST,Yan2018DeepLesion}. Each study typically contains multiple image volumes (series) that are scanned at the same time point but differ in reconstruction filters, contrast phases, etc. One lesion instance exists across multiple studies and series, but radiologists often do not mark them all in their daily work~\cite{Yan2018graph}. Besides, a large lesion spans in multiple slices in a volume, but radiologists generally mark it only on the slice where it has the largest cross-sectional size~\cite{Eisenhauer2009RECIST}, known as the key slice. These clinical prior knowledge gives us a chance to find those missing annotations that belong to the same lesion instance with existing annotations but were not marked by radiologists.

First, we train MELD using the existing annotations on key slices in the training set of DeepLesion. Then, we apply the network on all slices in the training set. After sampling a slice every 5mm, we obtained 1,429K proposals from 233K partially-labeled and unlabeled slices, a large extension compared to the 22K key slices. The next step is to establish correspondence between the proposals and existing annotations. We leverage the lesion embedding generated by LesaNet~\cite{Yan2019Lesa}, which encodes the body part, type, and attributes of lesions and have proved its efficacy in lesion retrieval. The distance of two embeddings should be small if they are from the same lesion instance. Hence, within each patient, we compute the L2 distance between every annotation and every proposal and keep those pairs whose distance is smaller than a threshold $ \theta $.

\Fig{MAM_example} illustrates three pairs of matched lesions. We found the mined lesions are mostly the same instances with existing ones, but sometimes they are actually different instances with similar semantic attributes, \eg, two liver metastatic tumors. Note that the mined lesions have similar but not identical appearance with existing ones, since they have different time point, reconstruction kernel, slice position, etc., see \Fig{MAM_example}. Therefore, the mined ones can still provide valuable new information when they are used in training. 
They are only used in training the classification heads but not the bounding-box regression head, since they are detected proposals and the box may be inaccurate.

\subsection{Negative Region Mining (NRM)}
\label{subsec:NRM}

Besides positive samples, negative samples are also important when training a detector. They are sampled from the background region of training images. If the background region contains missing annotations, the algorithm will learn from wrong supervision signals and degrade in accuracy \cite{Niitani2019sample,Wu2019soft}. In our problem, the MAM strategy cannot find all missing annotations because some of them do not belong to the same instance with any existing annotation. One idea is to treat all universal proposals of MELD as missing annotations. However, there may actually be many false positives (FPs) in the proposals.

Our solution is to explore the semantic overlap between datasets and seek help from the single-type datasets. Recall that MELD is an ensemble of four dataset experts, namely the DeepLesion expert and three single-type experts: LUNA, LiTS, and NIH-LN. For each slice in the training set of DeepLesion, it can output four groups of proposals. Compared to the universal proposals from the DeepLesion expert, the single-type proposals generally have fewer FPs in their specialties. This is because each single-type expert only needs to learn to detect a single lesion type, which is a much simpler task. Also, their training datasets are fully-labeled. For each proposal from the three single-type experts, if its detection score is higher than a threshold $ \sigma $ and it does not overlap with existing or mined annotations, we regard the proposal as a suspicious lesion. Then, we can either treat the suspicious lesions as positive samples or ignore them (do not sample them as either positive or negative) during finetuning. It is found that ignoring them achieved better accuracy, which prevents the FPs in these suspicious proposals from polluting the positive sample set. We also note that the suspicious lesions only include lung nodules, liver tumors, and LNs due to the single-type datasets used. Adding more single-type datasets will help to mine more suspicious lesions. 

Apart from the existing annotations, the mined missing annotations, and the suspicious lesions, the rest part of an image in DeepLesion is treated as reliable negative region, as depicted in \Fig{framework}. Previous ULD algorithms~\cite{Zlocha2019retina,Wang2019universal,Li2019MVP,Yan2019MULAN} were all limited to the 22K labeled training slices. It will bias the algorithms toward lesion-rich body parts and cause many FPs in under-represented body parts. With MAM and NRM, we can exploit the massive unlabeled slices and improve performance on the whole body. We anticipate the proposed methods to also be useful in other large-scale but partially-labeled datasets such as OpenImage~\cite{Kuznetsova2018open,Niitani2019sample}, where the missing annotations may be mined using embedding-based matching and with the help of other specialized object datasets.



\section{Experimental Details}
\label{sec:exp}

\subsection{Data}
\label{subsec:data}

To date, DeepLesion~\cite{Yan2018DeepLesion} is the largest dataset for universal lesion detection, containing 32,735 lesions annotated on 32,120 axial CT slices from 10,594 studies of 4,427 patients. It was mined from the National Institutes of Health Clinical Center based on marks annotated by radiologists during their routine work to measure significant image findings~\cite{Eisenhauer2009RECIST}. Thus, it closely reflects clinical needs. The LUNA (LUng Nodule Analysis) dataset~\cite{Setio2017LUNA} consists of 1,186 lung nodules annotated in 888 CT scans. LiTS (LIver Tumor Segmentation Benchmark)~\cite{Bilic2019LiTS} includes 201 CT scans with 0 to 75 liver tumors annotated per scan. We used 131 scans of them with released annotations. NIH-Lymph Node (NIH-LN)~\cite{NIH_LN_dataset} contains 388 mediastinal LNs in 90 CT scans and 595 abdominal LNs in 86 scans. Without loss of generality, we chose these three single-type datasets for joint learning with DeepLesion. Single-type datasets of other organs can be added in the future. More dataset details will be described in the supplementary material.

For DeepLesion, we used the official data split which has 70\%, 15\%, 15\% for training, validation, and test, respectively. The official test set includes only key slices and may contain missing annotations, which will bias the accuracy. We invited a board-certified radiologist to further fully annotate 1071 sub-volumes chosen from the test set of DeepLesion using the same RECIST criterion~\cite{Eisenhauer2009RECIST} as in DeepLesion. We call the official test set ``key-slice test set'' and the new one ``volumetric test set''. In the latter set, there are 1,642 original annotations and 2,023 manually added ones. For LUNA, LiTS, and NIH-LN, we randomly used 80\% of each dataset for joint training with DeepLesion, and left 20\% for validation. Image preprocessing and data augmentation steps are the same with~\cite{Yan2019MULAN}, which will be described in detail in the supplementary material.

\subsection{Implementation}
\label{subsec:implementation}

The proposed framework was implemented in PyTorch based on the maskrcnn-benchmark project~\cite{massa2018mrcnn}. The backbone of MELD is DenseNet-121~\cite{Huang2017DenseNet} initialized with an ImageNet pretrained model. The gating head has two FC-512 (fully-connected layers with 512 neurons), one FC-3 (3 organs), and a sigmoid function. A classification head consists of two FC-1024 
followed by an FC-4 (4 datasets). We use lung, liver, and LN as organ experts since they are the most common organs in DeepLesion~\cite{Yan2019Lesa}. These layers were randomly initialized. Each mini-batch had 4 samples, where each sample consisted of 9 axial CT slices for 3D feature fusion~\cite{Yan2019MULAN}. We used Rectified Adam (RAdam)~\cite{liu2019radam} to train MELD for 8 epochs and set the base learning rate to 0.0001, then reduced it by a factor of 10 after the 4th and 6th epochs. For single-type datasets, we used all slices that contain lesions and the same number of randomly sampled negative slices (without lesions) to train in each epoch. It took MELD 35ms to process a slice during inference on a Quadro RTX 6000 GPU. 

For MAM, we empirically set the distance threshold $ \theta=0.15 $. 27K missing annotations were mined from the training set of DeepLesion, in addition to the 23K existing annotations. We randomly checked 100 of them and found 90\% are true lesions. For NRM, we set the detection score threshold $ \sigma=0.5 $. An average of 0.45 suspicious lesions were detected per slice. We then finetuned MELD from an intermediate checkpoint in the 4th epoch with RAdam for 4 epochs using the same learning rate schedule ($ 10^{-5} $ to $ 10^{-6} $). In each finetuning epoch, we kept the original 22K key slices and randomly selected 10K unlabeled slices to add into the training set. MAM and NRM were used to mine missing annotations and reliable negative region in these 32K slices.

\subsection{Metrics}
\label{subsec:metrics}

The free-response receiver operating characteristic (FROC) curve is the standard metric in lesion detection~\cite{Setio2017LUNA,Yan2018DeepLesion,Shin2016TMICNN}. Sensitivities at different number of FPs per image are calculated to show the recall at different precision levels. We evaluate the ULD accuracy on the fully-annotated volumetric test set of DeepLesion. Following the LUNA challenge~\cite{Setio2017LUNA}, sensitivities at $1/8, 1/4, 1/2, 1, 2, 4, 8$ FPs per sub-volume are computed. Note that our 2.5D framework outputs 2D detections per slice, while this metric is for 3D detections. Thus, we designed a simple heuristic approach to stack 2D boxes to 3D ones if the intersection over union (IoU) of two 2D boxes in consecutive slices is greater than 0.5. If any 2D cross-section of a stacked 3D box has an IoU $ >0.5 $ with a 2D ground-truth box, the 3D box is counted as a TP. To compare with prior work, we also calculated the sensitivities at 0.5, 1, 2, and 4 FPs per key slice on the key-slice test set.

\section{Results and Discussion}
\label{sec:res}


In this section, we evaluate the effectiveness of our proposed algorithms: multi-expert lesion detector (MELD), missing annotation matching (MAM), and negative region mining (NRM). Our baseline method is the previous state of the art on DeepLesion, MULAN~\cite{Yan2019MULAN}. In \Table{ablation}, we can find that MELD outperformed the baseline by 1.7\% in average sensitivity at different FP levels. Adding MAM and NRM both significantly boosted the accuracy. This means that the missing annotations play a critical role in the detector's performance. MAM added matched lesions to the positive sample set to make the algorithm learn more about the appearance of different lesions. NRM removed suspicious lesions from the negative sample set to reduce its noise, so that the algorithm can learn the appearance of normal tissues better. Finally, MELD with both MAM and NRM achieved the best result, a relative improvement of 29\% compared to the baseline. We also explored to add different single-type datasets to mine suspicious lesions in NRM. \Table{NRM_organ} listed the detection accuracy of lesions in different organs. We can see that adding a dataset is generally beneficial for lesions in the corresponding organ, confirming the effectiveness of our algorithm to transfer knowledge from single-type datasets.

\begin{table}[t]
	\begin{center}
		\setlength{\tabcolsep}{4pt}
		\caption{Results with different components of the proposed framework . Sensitivity (\%) at different FPs per sub-volume on the volumetric test set of DeepLesion is shown.}
		\label{tbl:ablation}
		\begin{tabular}{lcccccccc}
			\hline\noalign{\smallskip}
			Method	& FP@0.125	& 0.25	& 0.5	& 1	& 2	& 4	& 8 & Average \\ 
			\noalign{\smallskip}\hline\noalign{\smallskip}
			Baseline~\cite{Yan2019MULAN}	& ~7.6	& 12.6	& 20.7	& 30.6	& 42.1	& 51.8	& 61.2	& 32.4	\\
			MELD	& ~7.7	& 13.5	& 21.3	& 32.3	& 43.7	& 54.8	& 65.2	& 34.1	\\
			MELD+MAM	& 12.9	& 20.8	& 29.2	& 39.0	& 49.6	& 58.7	& 67.0	& 39.6 	\\
			MELD+NRM	& 13.5	& 20.2	& 29.5	& 38.8	& 49.4	& 58.5	& 67.3	& 39.6 	\\
			MELD+MAM+NRM	& \bf 16.0	& \bf 22.8	& \bf 32.0	& \bf 41.7	& \bf 51.3	& \bf 60.3	& \bf 68.3	& \bf 41.8 	\\
			\hline
		\end{tabular}
	\end{center}
\end{table}

\begin{table}[t]
	\begin{center}
		\setlength{\tabcolsep}{4pt}
		\caption{Organ-stratified results with different single-type datasets used. In each row, we use certain datasets in NRM, then compute the detection accuracy of lesions in different organs. Bold results indicate the best accuracy for each organ (column). Average sensitivity at FP=0.125$ \sim $8 per sub-volume on DeepLesion is shown.}
		\label{tbl:NRM_organ}
		\begin{tabular}{lcccc}
			\hline\noalign{\smallskip}
			Single-type dataset	& Lung	& Liver	& Lymph node & Overall	 \\ 
			\noalign{\smallskip}\hline\noalign{\smallskip}
			LUNA (lung nodules)	& \bf 35.5	& 31.7	& 31.7	& 39.4	\\
			LiTS (liver tumors)	& 34.6	& \bf 39.1	& 33.8	& 40.1	\\
			NIH-LN (lymph nodes)	& 34.0	& 31.8	& 33.0	& 39.6	\\
			All		& 35.1	& 38.9	& \bf 35.3	& \bf 41.8	\\
			\hline
		\end{tabular} \vspace{-3mm}
	\end{center}
\end{table}

\begin{figure}[]
	\centering
	\includegraphics[width=.9\linewidth,trim=0 00 0 0, clip]{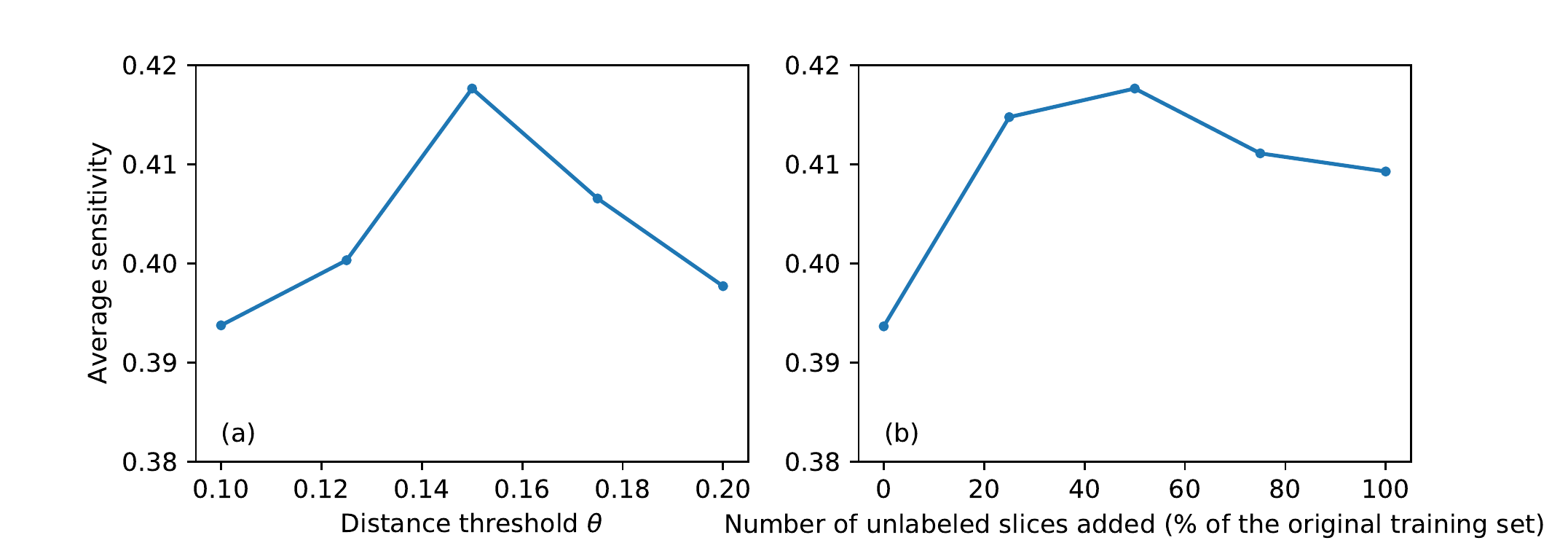} 
	\caption{Parameter study of the proposed algorithms. Average sensitivity at FP=0.125$ \sim $8 per sub-volume on DeepLesion is shown. In (b), the $ x $-axis is the ratio between the number of added slices and the original training size (22K key slices).}
	\label{fig:param_study}
\end{figure}

The influence of different parameter values is studied in \Fig{param_study}. In MAM, if the distance threshold $ \theta $ is too small, fewer missing annotations will be matched, providing less new information; If it is too large, the matched missing annotations may be noisy. Sub-plot (b) shows that adding unlabeled training images is helpful. With MAM and NRM, the accuracy was already improved on the original training set with no added slices (from MELD's 34.1\% in \Table{ablation} to 39.4\%). With more unlabeled slices added, MAM and NRM can find positive and negative samples that bring new information, 
especially for under-represented body parts in the original training set. The accuracy reached the best when the number of added slices is about half of the size of the original training set.

\begin{figure}[t]
	\centering
	\includegraphics[width=.5\linewidth,trim=10 0 20 30, clip]{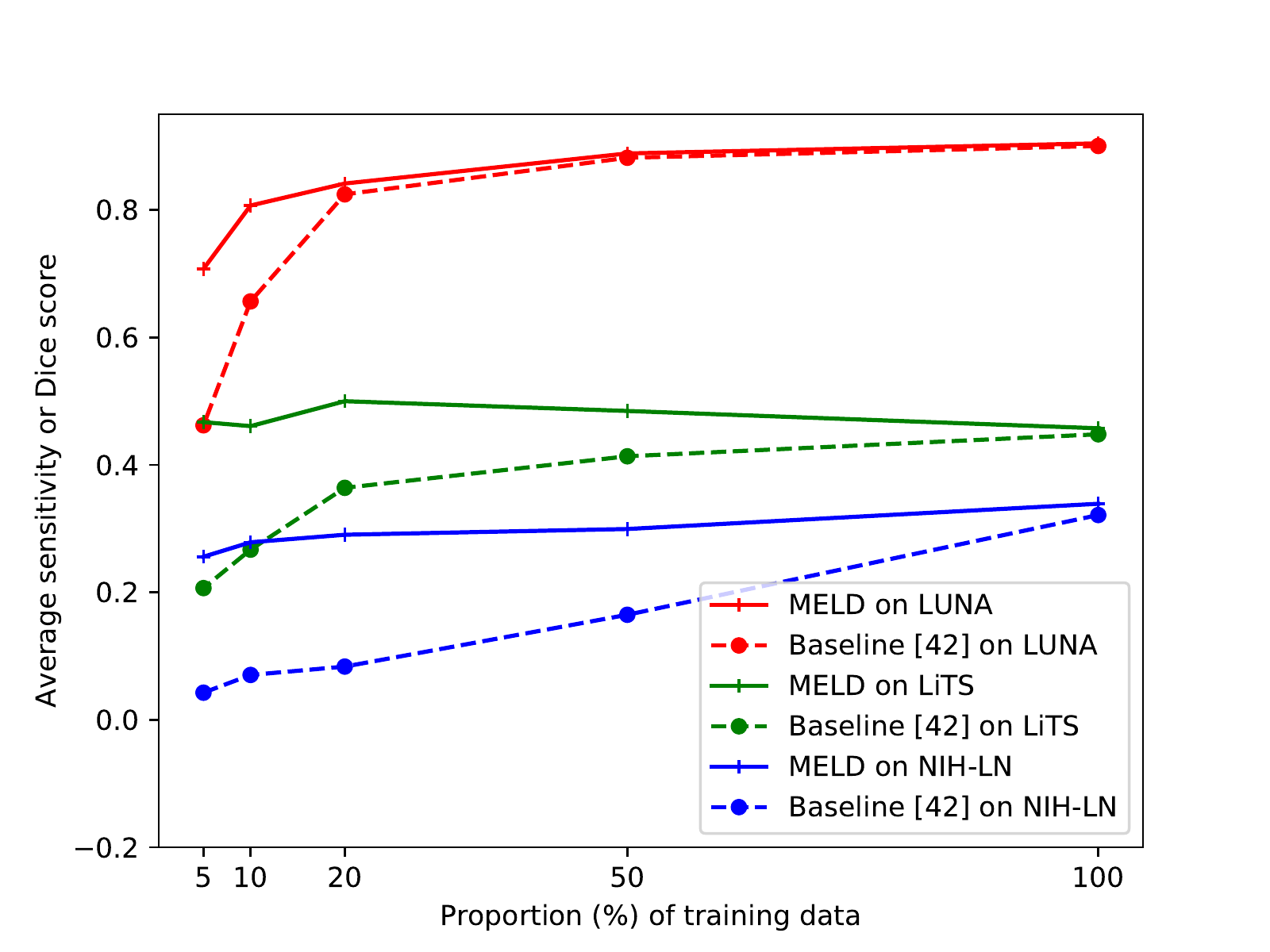} 
	\caption{Comparison of the baseline~\cite{Yan2019MULAN} and MELD with different proportions of training data in the single-type datasets. On LUNA, we report the average sensitivity at $ 1/8\sim 8 $ FPs per volume~\cite{Setio2017LUNA}. On LiTS and NIH-LN which have ground-truth masks, we report the Dice score.}
	\label{fig:spec_acc} \vspace{-5mm}
\end{figure}

The joint training strategy in MELD can improve the baseline not only on DeepLesion, but also on single-type datasets, especially when the number of training samples is small. Note that it is not our goal to compare with best algorithms specially designed for each single-type dataset. We combined DeepLesion with a proportion of training volumes from all single-type datasets to train MELD. For comparison, we trained the baseline~\cite{Yan2019MULAN} with one single-type dataset each time of the same training size. Evaluation was made on the validation set (20\% of each dataset). \Fig{spec_acc} shows that MELD always outperformed the baseline on the three single-type datasets. MELD's superiority is more evident when the number of training data is getting smaller. This is because DeepLesion contains lesions in a variety of organs, so it can help the single-type datasets learn effective features in the network backbone and organ heads. It is especially useful in medical image analysis where training data is often limited. It also indicates that the network has the capacity to learn different lesion types in multiple datasets at the same time. Among the three single-type datasets, lung nodules have relatively distinct appearance (\Fig{dataset_examples}), thus are easier to learn. Besides, LUNA has the more training data, so the superiority of MELD is smaller. Some liver tumors have clear separation with normal tissues, while others can be subtle, making it a harder task. Lymph nodes exist throughout the body and are sometimes hard to be discriminated from the surrounding vessels, muscles, and other organs, leading to the lowest accuracy.

\subsection{Comparison with Other Methods}
\label{subsec:comparison}

Previous works were evaluated on the partially-labeled key-slice test set. We use the same criterion and compare with other methods in \Table{prior_art}. MELD outperformed the previous state-of-the-art method, MULAN, either without or with the extra training information of 171 lesion tags~\cite{Yan2019MULAN}. MAM and NRM further boosted the accuracy and demonstrated that the mined missing annotations and reliable negative regions are helpful.

\begin{table}[t]
	\begin{center}
		\setlength{\tabcolsep}{4pt}
		\caption{Comparison with previous studies. Sensitivity (\%) at different FPs per \textit{slice} on the key-slice test set of DeepLesion is shown.}
		\label{tbl:prior_art}
		\begin{tabular}{lccccc}
			\hline\noalign{\smallskip}
			Method	& FP@0.5	& 1	& 2	& 4	& Average \\
			\noalign{\smallskip}\hline\noalign{\smallskip}
			ULDor~\cite{Tang2019Uldor}	& 52.9	& 64.8	& 74.8	& 84.4	& 69.2	\\
			Domain-attentive universal detector~\cite{Wang2019universal}	& -	& -	& -	& 87.3	& -	\\
			Volumetric attention~\cite{Wang2019LiTS}	& 69.1	& 77.9	& 83.8	& 87.5	& 79.6 \\
			MVP-Net~\cite{Li2019MVP}	& 73.8	& 81.8	& 87.6	& 91.3	& 83.6 \\
			MULAN (without tags)~\cite{Yan2019MULAN}	& 76.1	&  82.5	& 87.5	& 90.9	& 84.3	\\
			MULAN (with 171 tags)~\cite{Yan2019MULAN}	& 76.1  & 83.7  & 88.8  & 92.3  & 85.2	\\
			MELD (proposed)				& 77.8	& 84.8	& 89.0	& 91.8	& 85.9	\\
			MELD+MAM+NRM (proposed)		& \bf 78.6	& \bf 85.5	& \bf 89.6	& \bf 92.5	& \bf 86.6	\\
			\hline
		\end{tabular}
		
	\end{center}
\end{table}

\begin{table}[t]
	\begin{center}
		\setlength{\tabcolsep}{4pt}
		\caption{Different strategies to combine multiple datasets. Average sensitivity (\%) at FP=0.125$ \sim $8 per sub-volume on the volumetric test set of DeepLesion is shown.}
		\label{tbl:multi_dataset}
		\begin{tabular}{lc}
			\hline\noalign{\smallskip}
			Method	&  Average sensitivity \\
			\noalign{\smallskip}\hline\noalign{\smallskip}
			Single dataset (Baseline~\cite{Yan2019MULAN})	& 32.4 \\
			Dataset concatenation	& 34.5 \\
			Dataset concatenation (positive only)	& 33.1 \\
			Multi-task learning	& 35.1 \\
			Proposed (suspicious lesions as positive)	& 40.7 \\
			Proposed (suspicious lesions as ignore)	& \bf 41.8 \\
			\hline
		\end{tabular} \vspace{-5mm}
	\end{center}
\end{table}

Different strategies to combine multiple lesion datasets are compared in \Table{multi_dataset}. The baseline~\cite{Yan2019MULAN} is trained using a single dataset (DeepLesion). A straightforward way to incorporate the single-type datasets is to directly concatenate them with DeepLesion and treat them as one task. Another method is to only sample positive regions from them to avoid the influence of unannotated lesions of other types. These two methods combine multiple datasets in the data level and slightly improved the baseline. Because the datasets have heterogeneous labels, a better solution is to use multi-task learning and treat each dataset as a task and let them learn shared features. We proposed negative region mining to leverage single-type datasets in a novel way: using them to mine suspicious lesions in a universal dataset, which is similar to knowledge distillation~\cite{Hinton2014distill}. We also find that treating the mined lesions as ignore is better than regarding them as true lesions, possibly because they contain some noise and concept shift.

\subsection{Qualitative Results}
\label{subsec:qualitative}

\begin{figure}[t]
	\centering
	\includegraphics[width=\linewidth,trim=0 100 70 0, clip]{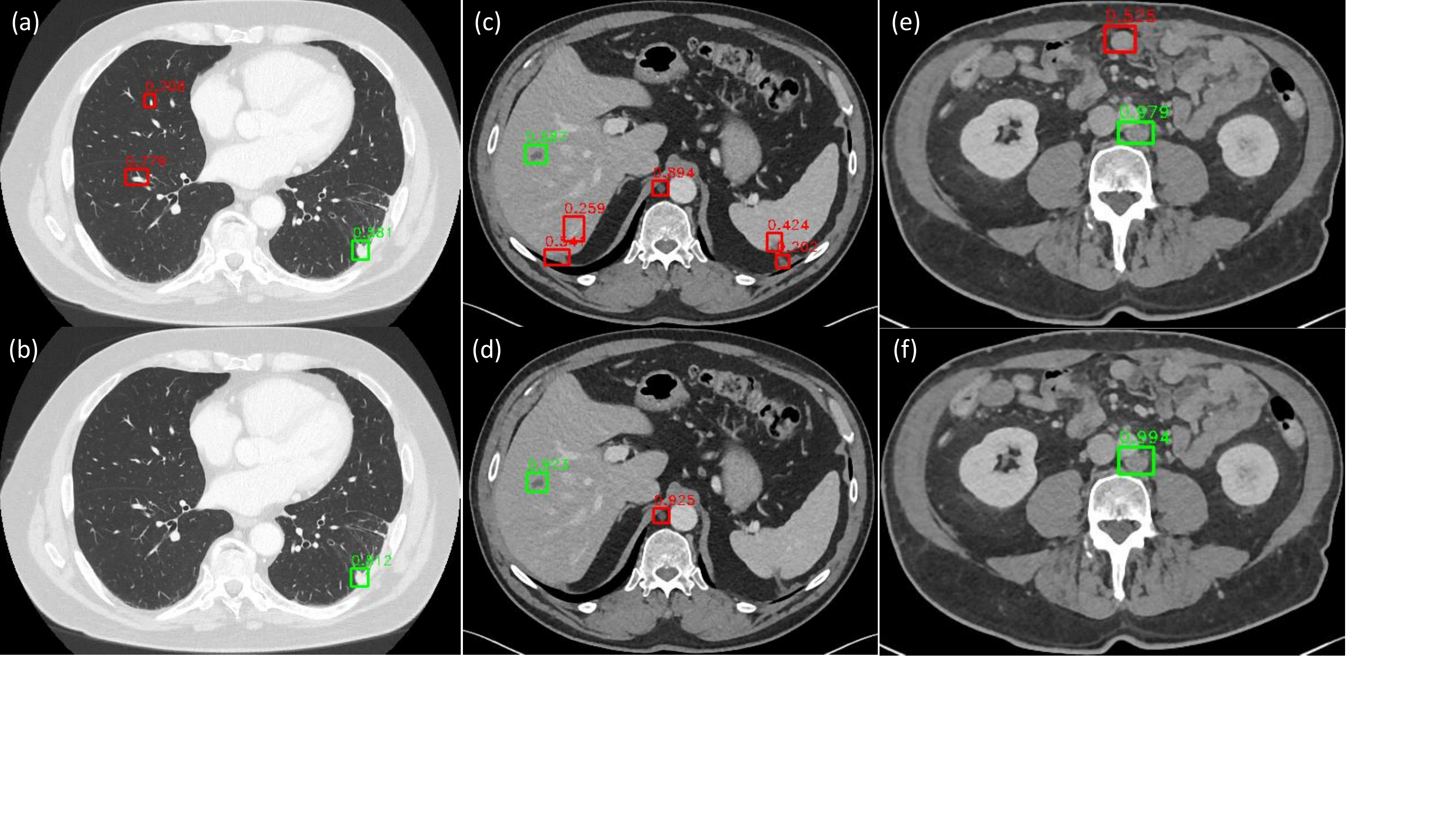} 
	\caption{Comparison of the baseline~\cite{Yan2019MULAN} (top row) and our proposed method (bottom row) on the volumetric test set of DeepLesion. Green and red boxes indicate TPs and FPs, respectively. Numbers above boxes are detection scores.}
	\label{fig:det_res} \vspace{-3mm}
\end{figure}

Qualitative comparison of the baseline~\cite{Yan2019MULAN} and the proposed method is shown in \Fig{det_res}. The baseline mistakenly detected vessels in the lung and liver, diaphragm, and bowels because of their similar appearance with lesions. These FPs have been reduced notably by our MELD+MAM+NRM. From \Table{ablation} we can find that the baseline achieved 50\% recall at 4 FPs per sub-volume while the proposed method achieved it at 2 FPs. On the other hand, the scores of TPs also increased in our method. This is because MAM added more positive samples and NRM excluded many missing annotations when finetuning MELD. Without the wrong negative training signals, true lesions can be learned more confidently.  

\section{Conclusion}
\label{sec:conclusion}

In this paper, we proposed a framework to mine positive and negative regions from a partially-labeled dataset for lesion detection. A multi-expert lesion detector with organ stratification was proposed to generate lesion proposals. Medical knowledge was leveraged to find missing annotations with an embedding matching approach. Multiple single-type datasets were utilized to mine suspicious lesions and generate reliable negative regions, so as to transfer their knowledge to the universal lesion detection model. As a result, our framework provides a powerful means to exploit multi-source, heterogeneously and imperfectly labeled data, significantly pushing forward universal lesion detection performance.

\section{Appendix}
\subsection{More Details on Datasets}
\label{subsec:dataset}

\begin{table}[]
	\begin{center}
		\scriptsize
		\setlength{\tabcolsep}{4pt}
		\renewcommand{\arraystretch}{1.5}
		\caption{Statistics of the four lesion datasets used in our work.}
		\label{tbl:datasets}
		\begin{tabular}{p{2.5cm}p{2cm}p{1cm}p{1.2cm}p{.9cm}p{1cm}p{1cm}}
            \hline\hline
			Name	& Lesion types	& Organs	& \# 3D Volumes	& \# 2D Slices	& \# Lesions	& Fully-annotated?	 \\
            \hline\hline
			DeepLesion~\cite{Yan2018DeepLesion}	& Various	& Whole body	& 10,594 sub-volumes	& 928K	& 32,735	& No \\
			\hline
			LUNA (LUng Nodule Analysis)~\cite{Setio2017LUNA}	& Lung nodule	& Lung	& 888	& 226K	& 1,186	& Yes \\
			\hline
			LiTS (LIver Tumor Segmentation Benchmark)~\cite{Bilic2019LiTS}	& Liver tumor	& Liver	& 130	& 85K	& 908	& Yes \\
			\hline
			NIH-LN (NIH-Lymph Node)~\cite{NIH_LN_dataset}	& Mediastinal and abdominal lymph nodes	& Lymph node	& 176	& 134K	& 983	& Yes \\			\hline\hline
		\end{tabular}
	\end{center}
\end{table}

The four lesion datasets used in our work are summarized in \Table{datasets}.

\begin{figure}[]
	\begin{center}
		\includegraphics[width=.8\linewidth,trim=10 40 230 30, clip]{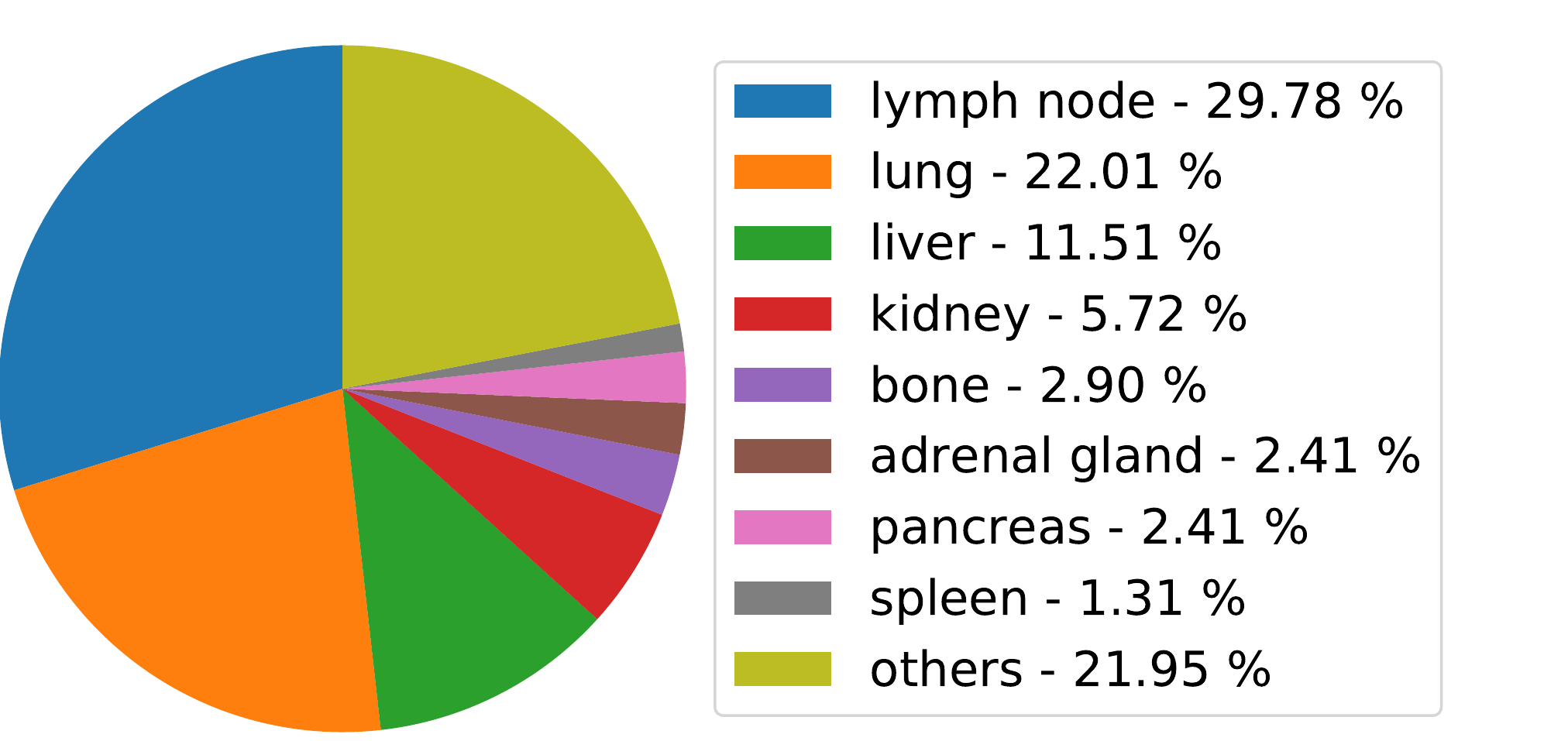} 
	\end{center}
	\caption{Statistics of major organs of lesions in DeepLesion~\cite{Yan2019Lesa}.}
	\label{fig:tag_cnt}
\end{figure}

\Fig{tag_cnt} shows the statistics of major organs of lesions in DeepLesion~\cite{Yan2018DeepLesion}. Based on the lesion tags provided by~\cite{Yan2019Lesa}, we analyzed 17,705 lesions with body part tags. Lymph node (LN), lung, and liver are the most common organs, which are covered by the organ heads in our multi-expert lesion detector (MELD) and our chosen single-type lesion datasets~\cite{Setio2017LUNA,Bilic2019LiTS,NIH_LN_dataset}. Note that it is easy to extend our proposed MELD and negative region mining (NRM) to more organs heads and more lesion datasets (\eg, tumors in kidney~\cite{KITS_dataset}, pancreas, colon~\cite{simpson2019large}, etc.).

\begin{figure}[]
	\begin{center}
		\includegraphics[width=.8\linewidth,trim=30 20 30 40, clip]{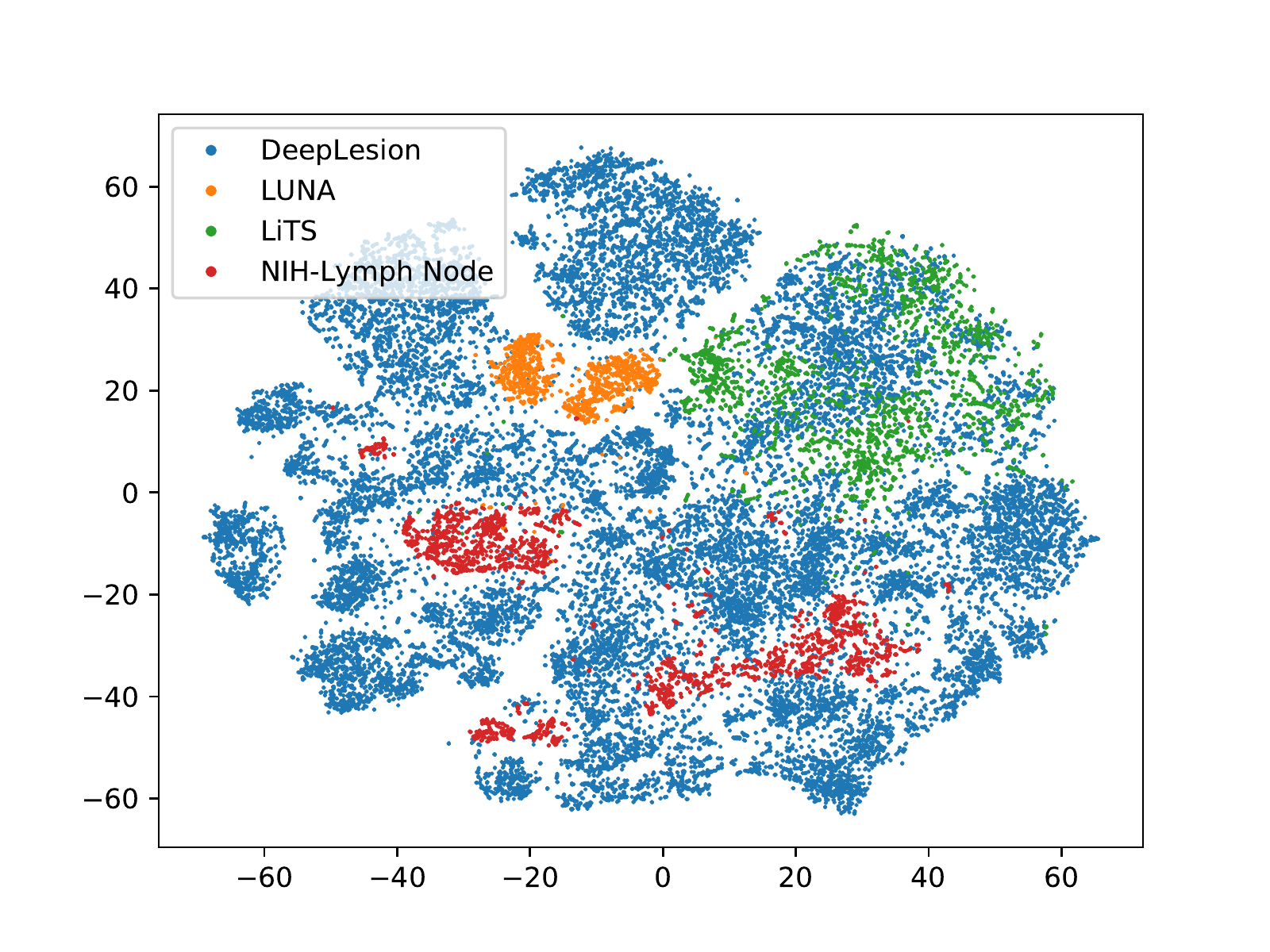} 
	\end{center}
	\caption{Scatter map of embeddings~\cite{Yan2019Lesa} of lesions in DeepLesion~\cite{Yan2018DeepLesion}, LUNA~\cite{Setio2017LUNA}, LiTS~\cite{Bilic2019LiTS}, and NIH-LN~\cite{NIH_LN_dataset} computed by t-SNE.}
	\label{fig:dataset_tsne}
\end{figure}

To observe the distribution of the four datasets, we calculated the 256D lesion embeddings from LesaNet~\cite{Yan2019Lesa} and visualize them using t-SNE~\cite{Maaten2014tsne}.  From \Fig{dataset_tsne}, we can find the single-type datasets lie within subspaces of DeepLesion. NIH-LN is more scattered as lymph nodes exist throughout the body and have diverse contextual appearance. 

\subsection{More Details on Preprocessing}
\label{subsec:impl_detail}

The image preprocessing and data augmentation steps in our experiments are described in this section. We tried to build a unified lesion detection/segmentation framework for various datasets, and used the same workflow for all input images. First, we normalized the orientations of $ x,y,$ and $z $-axes of all datasets to the same direction. Then, we rescaled the 12-bit CT intensity range to floating-point numbers in [0,255] using a single windowing (-1024--3071 HU) that covers the intensity ranges of the lung, soft tissue, and bone. Every axial slice was resized so that each pixel corresponds to 0.8mm. We interpolated in the $ z $-axis to make the slice intervals of all volumes to be 2mm. The black borders in images were clipped for computation efficiency. When training, we did data augmentation by randomly resizing each slice with a ratio of 0.8$ \sim $1.2 and randomly shifting the image and annotation by -8$ \sim $8 pixels in $ x $ and $ y $ axes.

\clearpage
%
%
\bibliographystyle{splncs04}
\bibliography{lesion_det}
\end{document}